\documentclass{article}

\usepackage{graphicx}
\usepackage[normalem]{ulem}
\usepackage{booktabs}
\usepackage{subcaption}
\usepackage{amsmath,amsfonts,amssymb,amsthm}
\usepackage{mathtools}
\usepackage[dvipsnames]{xcolor}
\usepackage{multirow}
\usepackage{caption}
\usepackage[algo2e,vlined,ruled]{algorithm2e}
\usepackage[draft,layout={inline,index}]{fixme}
\usepackage{natbib}
\usepackage{pdfpages}
\usepackage{xr}

\def\appendixpath{rats-appendix/rats-appendix}
\def\includeappendixatbottom{1} 
\def\statecodewillbeprovided{1} 
\def\submissionoption{final} 
\usepackage[\submissionoption]{neurips_2019}
\usepackage[utf8]{inputenc} 
\usepackage[T1]{fontenc}    
\usepackage{hyperref}       
\usepackage{url}            
\usepackage{booktabs}       
\usepackage{amsfonts}       
\usepackage{nicefrac}       
\usepackage{microtype}      

\usepackage{notations}
\def\RATSmeaning{Risk-Averse Tree-Search}
\def\RATS{RATS}

\newtheorem{definition}{Definition}

\newtheorem{property}{Property}

\title{
	Non-Stationary Markov Decision Processes\\a Worst-Case Approach using Model-Based Reinforcement Learning
}
\author{%
	Erwan Lecarpentier\\
	Universit\'e de Toulouse\\
	ONERA - The French Aerospace Lab\\
	\texttt{erwan.lecarpentier@isae-supaero.fr} \\
	\And
	Emmanuel Rachelson \\
	Universit\'e de Toulouse\\
	ISAE-SUPAERO\\
	\texttt{emmanuel.rachelson@isae-supaero.fr}
}

\begin{document}
\maketitle

\begin{abstract}
	This work tackles the problem of robust planning in non-stationary stochastic environments. We study Markov Decision Processes (MDPs) evolving over time and consider Model-Based Reinforcement Learning algorithms in this setting. We make two hypotheses: 1) the environment evolves continuously with a bounded evolution rate; 2) a current model is known at each decision epoch but not its evolution. Our contribution can be presented in four points. 1) we define a specific class of MDPs that we call Non-Stationary MDPs (NSMDPs). We introduce the notion of regular evolution by making an hypothesis of Lipschitz-Continuity on the transition and reward functions w.r.t. time; 2) we consider a planning agent using the current model of the environment but unaware of its future evolution. This leads us to consider a worst-case method where the environment is seen as an adversarial agent; 3) following this approach, we propose the Risk-Averse Tree-Search (RATS) algorithm, a Model-Based method similar to minimax search; 4) we illustrate the benefits brought by RATS empirically and compare its performance with reference Model-Based algorithms.
\end{abstract}

\section{Introduction}
\label{sec:intro}

%
%
One of the hot topics of modern Artificial Intelligence (AI) is the ability for an agent to adapt its behavior to changing tasks.
In the literature, this problem is often linked to the setting of Lifelong Reinforcement Learning (LRL) \citep{silver2013lifelong,abel2018state,abel2018policy} and learning in non-stationary environments \citep{choi1999hidden,jaulmes2005learning,hadoux2015markovian}.
In LRL, the tasks presented to the agent change sequentially at discrete transition epochs \citep{silver2013lifelong}.
Similarly, the non-stationary environments considered in the literature often evolve abruptly \citep{hadoux2015markovian,hadoux2014sequential,doya2002multiple,da2006dealing,choi1999hidden,choi2000hidden,choi2001solving,campo1991state,wiering2001reinforcement}.
%
%
In this paper, we investigate environments continuously changing over time that we call \NSMDPmeaning{}es (\NSMDP{}s).
In this setting, it is realistic to bound the evolution rate of the environment using a Lipschitz Continuity (LC) assumption.

%
%
Model-based Reinforcement Learning approaches~\citep{sutton1998reinforcement} benefit from the knowledge of a model allowing them to reach impressive performances, as demonstrated by the Monte Carlo Tree Search (MCTS) algorithm~\citep{silver2016mastering}.
In this matter, the necessity to have access to a model is a great concern of AI~\citep{asadi2018lipschitz,jaulmes2005learning,doya2002multiple,da2006dealing}.
%
%
In the context of \NSMDP{}s, we assume that an agent is provided with a \emph{snapshot} model when its action is computed.
By this, we mean that it only has access to the current model of the environment but not its future evolution, as if it took a photograph but would be unable to predict how it is going to evolve.
This hypothesis is realistic, because many environments have a tractable state while their future evolution is hard to predict~\citep{da2006dealing,wiering2001reinforcement}.
In order to solve LC-\NSMDP{}s, we propose a method that considers the worst-case possible evolution of the model and performs planning \wrt{} this model.
This is equivalent to considering Nature as an adversarial agent. 
%
%
The paper is organized as follows:
first we describe the \NSMDP{} setting and the regularity assumption (Section~\ref{sec:nsmdp});
then we outline related works (Section~\ref{sec:related});
follows the explanation of the worst-case approach proposed in this paper (Section~\ref{sec:worst});
then we describe an algorithm reflecting this approach (Section~\ref{sec:rats});
finally we illustrate its behavior empirically (Section~\ref{sec:expe}).

\section{\NSMDPmeaning{}es}
\label{sec:nsmdp}

To define a \NSMDPmeaning{} (\NSMDP{}), we revert to the initial MDP model introduced by \citet{puterman2014markov}
, where the transition and reward functions depend on time.
\begin{definition}
	\label{def:nsmdp}
	\textbf{\NSMDP{}.}
	An \NSMDP{} is an MDP whose transition and reward functions depend on the decision epoch.
	It is defined by a 5-tuple $\{ \S, \T, \A, \left(p_t\right)_{t\in \T}, \left(r_t\right)_{t\in \T} \}$ where
	$\S$ is a state space;
	$\T \equiv \{ 1, 2, \ldots, N \}$ is the set of decision epochs with $N \leq +\infty$;
	$\A$ is an action space;
	$p_t(s' \mid s, a)$ is the probability of reaching state $s'$ while performing action $a$ at decision epoch $t$ in state $s$;
	$r_t(s, a, s')$ is the scalar reward associated to the transition from $s$ to $s'$ with action $a$ at decision epoch $t$.
\end{definition}
%
%
This definition can be viewed as that of a stationary MDP whose state space has been enhanced with time.
While this addition is trivial in episodic tasks where an agent is given the opportunity to interact several times with the same MDP, it is different when the experience is unique.
Indeed, no exploration is allowed along the temporal axis.
%
%
%
%
Within a stationary, infinite-horizon MDP with a discounted criterion, it is proven that there exists a Markovian deterministic stationary policy~\citep{puterman2014markov}.
It is not the case within \NSMDP{}s where the optimal policy is non-stationary in the most general case.
%
%
Additionally, we define the expected reward received when taking action $a$ at state $s$ and decision epoch $t$ as $R_t(s, a) = \E_{s' \sim p_t(\cdot \mid s, a)} \left[r_t(s, a, s') \right]$.
Without loss of generality, we assume the reward function to be bounded between $-1$ and $1$. 
In this paper, we consider discrete time decision processes with constant transition durations, which imply deterministic decision times in Definition~\ref{def:nsmdp}.
This assumption is mild since many discrete time sequential decision problems follow that assumption.
A non-stationary policy $\pi$ is a sequence of \emph{decision rules} $\pi_t$ which map states to actions (or distributions over actions).
For a stochastic non-stationary policy $\pi_t(a \mid s)$, the value of a state $s$ at decision epoch $t$ within an infinite horizon \NSMDP{} is defined, with $\gamma \in [0, 1)$ a discount factor, by:
\begin{equation*}
	V^\pi_t(s) = \E \left[ \sum_{i = t}^{\infty} \gamma^{i - t} R_i(s_i, a_i) \Big|
	\begin{aligned}
		&s_t = s, \ a_i \sim \pi_i(\cdot \mid s_i), \ s_{i+1} \sim p_i(\cdot \mid s_i, a_i)
	\end{aligned}
	\right],
\end{equation*}
The definition of the state-action value function $Q^\pi_t$ for $\pi$ at decision epoch $t$ is straightforward:
\begin{equation*}
	\label{eq:bellman}
	Q^\pi_t(s, a) = R_t(s, a) + \gamma \E_{s' \sim p_t(\cdot \mid s, a)} \left[ V^\pi_{t + 1}(s') \right].
\end{equation*}
Overall, we defined an \NSMDP{} as an MDP where we stress out the distinction between state, time, and decision epoch due to the inability for an agent to explore the temporal axis at will.
This distinction is particularly relevant for non-episodic tasks, \ie{} when there is no possibility to re-experience the same MDP starting from a prior date.

\textbf{The regularity hypothesis.}
Many real-world problems can be modeled as an \NSMDP{}.
For instance, the problem of path planning for a glider immersed in a non-stationary atmosphere~\citep{chung2015learning,lecarpentier2017empirical}, or that of vehicle routing in dynamic traffic congestion.
Realistically, we consider that the expected reward and transition functions do not evolve arbitrarily fast over time.
Conversely, if such an assumption was not made, a chaotic evolution of the \NSMDP{} would be allowed which is both unrealistic and hard to solve.
Hence, we assume that changes occur slowly over time.
Mathematically, we formalize this hypothesis by bounding the evolution rate of the transition and expected reward functions, using the notion of Lipschitz Continuity (LC).
\begin{definition}
	\textbf{Lipschitz Continuity.}
	Let $(X,d_X)$ and $(Y,d_Y)$ be two metric spaces and $f: X \rightarrow Y$, $f$ is $L$-Lipschitz Continuous ($L$-LC) with $L \in \R^+$ iff $d_Y(f(x), f(\hat{x})) \leq L \, d_X(x, \hat{x}) , \forall (x,\hat{x}) \in X^2$.
	$L$ is called a Lipschitz constant of the function $f$.
	\label{def:lc}
\end{definition}
We apply this hypothesis to the transition and reward functions of an \NSMDP{} so that those functions are LC \wrt{} time.
For the transition function, this leads to the consideration of a metric between probability density functions.
For that purpose, we use the 1-Wasserstein distance~\citep{villani2008optimal}.
\begin{definition}
	\label{def:wasserstein-distance}
	\textbf{1-Wasserstein distance.}
	Let $(X, d_X)$ be a Polish metric space, $\mu, \nu$ any probability measures on $X$, $\Pi(\mu, \nu)$ the set of joint distributions on $X \times X$ with marginals $\mu$ and $\nu$. The 1-Wasserstein distance between $\mu$ and $\nu$ is $W_1 (\mu, \nu) = \inf_{\pi \in \Pi(\mu, \nu)} \int_{X \times X} d_X(x,y) d\pi(x,y)$.
\end{definition}
The choice of the Wasserstein distance is motivated by the fact that it quantifies the distance between two distributions in a physical manner, respectful of the topology of the measured space \citep{dabney2018distributional,asadi2018lipschitz}.
First, it is sensitive to the difference between the supports of the distributions.
Comparatively, the Kullback-Leibler divergence 
between distributions with disjoint supports is infinite.
Secondly, 
if one consider two regions of the support where two distributions differ, the Wasserstein distance is sensitive to the distance between the elements of those regions.
Comparatively, the total-variation metric 
is the same regardless of this distance.
\begin{definition}
	\label{def:lc-nsmdp}
	\textbf{$(L_p, L_r)$-LC-\NSMDP{}.}
	An $(L_p, L_r)$-LC-\NSMDP{} is an \NSMDP{} whose transition and reward functions are respectively $L_p$-LC and $L_r$-LC \wrt{} time, \ie{}, $\forall (t, \hat{t}, s, s', a) \in \T^2 \times \S^2 \times \A$,
	\begin{equation*}
	W_1(p_t(\cdot \mid s, a), p_{\hat{t}}(\cdot \mid s, a)) \leq L_p |t - \hat{t}| \quad \textrm{and} \quad 
	|r_t(s, a, s') - r_{\hat{t}}(s, a, s')| \leq L_r |t - \hat{t}|.
	\end{equation*}
\end{definition}
One should remark that the LC property should be defined with respect to actual decision times and not decision epoch indexes for the sake of realism. In the present case, both have the same value, and we choose to keep this convention for clarity. Our results however extend easily to the case where indexes and times do not coincide.
From now on, we consider $(L_p, L_r)$-LC-\NSMDP{}s, making Lipschitz Continuity our regularity property.
Notice that $R$ is defined as a convex combination of $r$ by the probability measure $p$.
As a result, the notion of Lipschitz Continuity of $R$ is strongly related to that of $r$ and $p$ as showed by Property~\ref{prop:link-LR-LrLp}.
All the proofs of the paper can be found in the Appendix.
\begin{property}
	\label{prop:link-LR-LrLp}
	Given an $(L_p, L_r)$-LC-\NSMDP{}, the expected reward function $R_t: s, a \mapsto \E_{s' \sim p_t(\cdot \mid s, a)} \left\{ r_t(s, a, s') \right\}$ is $L_R$-LC with $L_R = L_r + L_p$.
\end{property}
This result shows $R$'s evolution rate is conditioned by the evolution rates of $r$ and $p$.
It allows to work either with the reward function $r$ or its expectation $R$, benefiting from the same LC property.

\section{Related work}
\label{sec:related}

\citet{iyengar2005robust} introduced the framework of robust MDPs, where the transition function is allowed to evolve within a set of functions due to uncertainty.
This differs from our work in two fundamental aspects:
1) we consider uncertainty in the reward model as well;
2) we use a stronger Lipschitz formulation on the set of possible transition and reward functions, this last point being motivated by its relevance to the non-stationary setting.
\citet{szita2002varepsilon} also consider the robust MDP setting and adopt a different constraint hypothesis on the set of possible functions than our LC assumption.
They control the total variation distance of transition functions from subsequent decision epochs by a scalar value.
Those slowly changing environments allow model-free RL algorithms such as Q-Learning to find near optimal policies.
\citet{lim2013reinforcement} consider learning in robust MDPs where the model evolves in an adversarial manner for a subset of $\S \times \A$.
In that setting, they propose to learn to what extent the adversary can modify the model and to deduce a behavior close to the minimax policy.
\citet{even2009online} studied the case of non-stationary reward functions with fixed transition models.
No assumption is made on the set of possible functions and they propose an algorithm achieving sub-linear regret w.r.t. the best stationary policy.
\citet{dick2014online} viewed a similar setting from the perspective of online linear optimization.
\citet{csaji2008value} studied the \NSMDP{} setting with an assumption of reward and transition functions varying in a neighborhood of a reference reward-transition function pair.
Finally, \citet{abbasi2013online} address the adversarial \NSMDP{} setting with a mixing assumption constraint instead of the LC assumption we make. 

Non-stationary environments also have been studied through the framework of Hidden Mode MDPs (\HMMDP{}) introduced by \citet{choi1999hidden}.
This is a special class of Partially Observable MDPs (POMDPs) \citep{kaelbling1998planning} where a hidden mode indexes a latent stationary MDP within which the agent evolves.
Similarly to the context of LRL, the agent experiences a series of different MDPs over time.
In this setting, \citet{choi1999hidden,choi2000hidden} proposed methods to learn the different models of the latent stationary MDPs. 
\citet{doya2002multiple} built a modular architecture switching between models and policies when a change is detected.
Similarly, \citet{wiering2001reinforcement,da2006dealing,hadoux2014sequential} proposed a method tracking the switching occurrence and re-planning if needed.
Overall, as in LRL, the \HMMDP{} setting considers abrupt evolution of the transition and reward functions whereas we consider a continuous one.
Other settings have been considered, as by \citet{jaulmes2005learning}, who do not make particular hypothesis on the evolution of the \NSMDP{}.
They build a learning algorithm for POMDPs solving, weighting recently experienced transitions more than older ones to account for the time dependency.

To plan robustly within an \NSMDP{}, our approach consists in exploiting the \emph{slow} LC evolution of the environment.
Utilizing Lipschitz continuity to infer bounds on a function is common in the RL, bandit and optimization communities \citep{kleinberg2008multi,rachelson2010locality,pirotta2015policy,pazis2013pac,munos2014bandits}.
We implement this approach with a minimax-like algorithm~\citep{fudenberg1991game}, where the environment is seen as an adversarial agent. 

\section{Worst-case approach}
\label{sec:worst}

We consider finding an optimal policy within an LC-\NSMDP{} under the non-episodic task hypothesis.
The latter prevents us from learning from previous experience data since they become outdated with time and no information samples have been collected yet for future time steps.
An alternative is to use model-based RL algorithms such as MCTS.
For a current state $s_0$, such algorithms focus on finding the optimal action $a_0^*$ by using a generative model.
This action is then undertaken and the operation repeated at the next state.
However, using the true \NSMDP{} model for this purpose is an unrealistic hypothesis, since this model is generally unknown.
We assume the agent does not have access to the true \NSMDP{} model; instead, we introduce the notion of \emph{snapshot model}.
Intuitively, the snapshot associated to time $t_0$ is a temporal slice of the \NSMDP{} at $t_0$.
\begin{definition}
	\textbf{Snapshot of an \NSMDP{}.} 
	The snapshot of an \NSMDP{} $\{ \S, \T, \A, \left(p_t\right)_{t\in \T}, \left(r_t\right)_{t\in \T} \}$ at decision epoch $t_0$, denoted by MDP$_{t_0}$, is the stationary MDP defined by the 4-tuple $\{\S, \A, p_{t_0}, r_{t_0}\}$ where
	$p_{t_0}(s' \mid s, a)$ and $r_{t_0}(s, a, s')$ are the transition and reward functions of the \NSMDP{} at $t_0$.
\end{definition}
Similarly to the \NSMDP{}, this definition induces the existence of the snapshot expected reward $R_{t_0}$ defined by $R_{t_0} : s, a \mapsto \E_{s'\sim p_{t_0}(\cdot \mid s, a)} \left\{ r_{t_0}(s, a, s') \right\}$.
Notice that the snapshot MDP$_{t_0}$ is stationary and coincides with the \NSMDP{} \emph{only} at $t_0$.
Particularly, one can generate a trajectory $\{s_0, r_0, \cdots, s_k\}$ within an \NSMDP{} using the sequence of snapshots $\{\text{MDP}_{t_0}, \cdots, \text{MDP}_{t_0 + k - 1}\}$ as a model.
Overall, the hypothesis of using snapshot models amounts to considering a planning agent only able to get the current stationary model of the environment.
In real-world problems, predictions often are uncertain or hard to perform \eg{} in the thermal soaring problem of a glider.

We consider a generic \emph{planning} agent at $s_0, t_0$, using MDP$_{t_0}$ as a model of the \NSMDP{}.
By planning, we mean conducting a look-ahead search within the possible trajectories starting from $s_0, t_0$ given a model of the environment.
The search allows in turn to identify an optimal action \wrt{} the model.
This action is then undertaken and the agent jumps to the next state where the operation is repeated.
The consequence of planning with MDP$_{t_0}$ is that the estimated value of an $s, t$ pair is the value of the optimal policy of $\text{MDP}_{t_0}$, written $V^*_{\text{MDP}_{t_0}}(s)$.
The true optimal value of $s$ at $t$ within the \NSMDP{} does not match this estimate because of the non-stationarity.
The intuition we develop is that, given the \emph{slow} evolution rate of the environment, for a state $s$ seen at a future decision epoch during the search, we can predict a scope into which the transition and reward functions at $s$ lie.
\begin{property}
	\textbf{Set of admissible snapshot models.}
	\label{prop:set-admissible-future-snapshot}
	Consider an $(L_p, L_r)$-LC-\NSMDP{}, $s, t, a \in \STA{}$.
	The transition and expected reward functions $(p_t, R_t)$ of the snapshot $\text{MDP}_t$ respect
	\begin{equation*}
	(p_t, R_t) \in \Delta_{t} \eqdef \B_{W_1} \left( p_{t - 1}(\cdot \mid s, a), L_p \right) \times \B_{| \cdot |} \left( R_{t - 1}(s, a), L_R \right)
	\end{equation*}
	where $L_R = L_p + L_r$ and $\mathcal{B}_d \left( c, r \right)$ denotes the ball of centre $c$, defined with metric $d$ and radius $r$.
\end{property}
For a future prediction at $s, t$, we consider the question of using a better model than $p_{t_0}, R_{t_0}$.
The underlying evolution of the \NSMDP{} being unknown, a desirable feature would be to use a model leading to a policy that is \emph{robust} to every possible evolution.
To that end, we propose to use the snapshots corresponding to the worst possible evolution scenario under the constraints of Property~\ref{prop:set-admissible-future-snapshot}.
We claim that such a practice is an efficient way to 1) ensure robust performance to all possible evolutions of the \NSMDP{} and 2) avoid catastrophic terminal states.
Practically, this boils down to using a different value estimate for $s$ at $t$ than $V^*_{\text{MDP}_{t_0}}(s)$ which provided no robustness guarantees.

Given a policy $\pi = \left(\pi_t\right)_{t\in \T}$ and a decision epoch $t$, a worst-case \NSMDP{} corresponds to a sequence of transition and reward models minimizing the expected value of applying $\pi$ in any pair $(s,t)$, while remaining within the bounds of Property~\ref{prop:set-admissible-future-snapshot}. We write $\overline{V}^{\pi}_t(s)$ this value for $s$ at decision epoch $t$.
\begin{equation}
	\label{eq:worst-case-v}
	\overline{V}^{\pi}_{t}(s) \eqdef \min_{
		\substack{\left( p_i, R_i \right) \in \Delta_i, \forall i \in \T}
	} \E \left[ \sum_{i = t}^{\infty} \gamma^{i-t} R_{i}(s_i, a_i)
	\Big|
	\begin{aligned}
		& s_t = s\\
		& a_i \sim \pi_i(\cdot \mid s_i), s_{i+1} \sim p_{i}(\cdot \mid s_i, a_i)
	\end{aligned} \right]
\end{equation}
Intuitively, the worst-case \NSMDP{} is a model of a non-stationary environment leading to the poorest possible performance for $\pi$, while being an admissible evolution of MDP$_t$.
Let us define $\overline{Q}^\pi_t(s,a)$ as the worst-case $Q$-value for the pair $(s,a)$ at decision epoch $t$:
\begin{equation}
	\label{eq:worst-case-q}
	\overline{Q}^\pi_t(s,a) \eqdef \min_{\substack{\left({p},R\right) \in \Delta_t}} \E_{s'\sim p}\left[R(s,a) + \gamma \overline{V}^{\pi}_{t+1}(s')\right].
\end{equation}

\section{\RATSmeaning{} algorithm}
\label{sec:rats}

\begin{figure}[]
	\begin{subfigure}{.4\textwidth}
		\centering
		\includegraphics[trim={2.2cm 0cm 1.9cm 0cm}, clip, width=\textwidth]{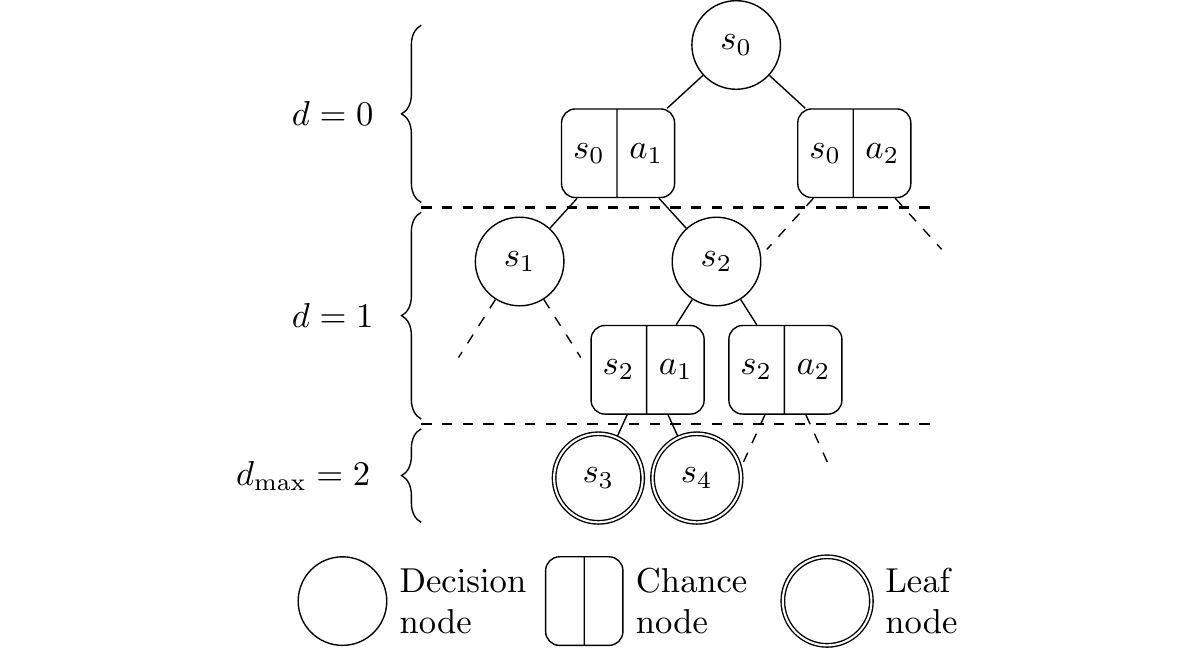}
		\captionsetup{width=\textwidth}
		\captionof{figure}{
			Tree structure, $d_{\max} = 2$, $\A = \{ a_1, a_2\}$.
		}
		\label{fig:tree}
	\end{subfigure}%
	\begin{subfigure}{.6\textwidth}
		\begin{tabular}{llccc}
			\toprule
			$\epsilon$ & & \RATS{} & DP-snapshot& DP-NSMDP\\
			\midrule
			
			\multirow{2}{4mm}{0}
			& $\E\left[\sum r\right]$ & -0.026 & \textbf{0.48} & \textbf{0.47}\\
			\cmidrule(r){2-5}
			& CVaR & \textbf{-0.81} & -0.90 & -0.9 \\
			\midrule
			
			\multirow{3}{4mm}{0.5}
			& $\E\left[\sum r\right]$ & \textbf{-0.032} & -0.46 & \textbf{-0.077} \\
			\cmidrule(r){2-5}
			& CVaR & \textbf{-0.81} & -0.90 & \textbf{-0.81} \\
			\midrule
			
			\multirow{3}{4mm}{1}
			& $\E\left[\sum r\right]$ & \textbf{0.67} & -0.78 & \textbf{0.66} \\
			\cmidrule(r){2-5}
			& CVaR & \textbf{0.095} & -0.90 & \textbf{-0.033}\\
			\bottomrule
		\end{tabular}
		\captionof{figure}{Expected return $\E\left[\sum r\right]$ and CVaR at 5\%.}
		\label{table:result}
	\end{subfigure}%
 	\caption{Tree structure and results from the  Non-Stationary bridge experiment.}
\end{figure}

\textbf{The algorithm.}
Tree search algorithms within MDPs have been well studied and cover two classes of search trees, namely closed loop~\citep{keller2013trial,kocsis2006bandit,browne2012survey} and open loop~\citep{bubeck2010open,lecarpentier2018open}.
Following \citep{keller2013trial}, we consider closed loop search trees, composed of \emph{decision nodes} alternating with \emph{chance nodes}.
We adapt their formulation to take time into account, resulting in the following definitions.
A decision node at depth $t$, denoted by $\nu^{s, t}$, is labeled by a unique state / decision epoch pair $(s, t)$.
The edges leading to its children chance nodes correspond to the available actions at $(s, t)$.
A chance node, denoted by $\nu^{s, t, a}$, is labeled by a state / decision epoch / action triplet $(s, t, a)$.
The edges leading to its children decision nodes correspond to the reachable state / decision epoch pairs $(s', t')$ after performing $a$ in $(s, t)$ as illustrated by Figure~\ref{fig:tree}.
We consider the problem of estimating the optimal action $a_0^*$ at $s_0, t_0$ within a worst-case \NSMDP{}, knowing MDP$_{t_0}$.
This problem is twofold. It requires 1) to estimate the worst-case \NSMDP{} given MDP$_{t_0}$ and 2)  to explore the latter in order to identify $a_0^*$.
We propose to tackle both problems with an algorithm inspired by the minimax algorithm~\citep{fudenberg1991game} where the \emph{max} operator corresponds to the agent's policy, seeking to maximize the return; and the \emph{min} operator corresponds to the worst-case model, seeking to minimize the return.
Estimating the worst-case \NSMDP{} requires to estimate the sequence of subsequent snapshots minimizing Equation~\ref{eq:worst-case-q}.
The inter-dependence of those snapshots (Equation~\ref{eq:worst-case-v}) makes the problem hard to solve \citep{iyengar2005robust},
particularly because of the combinatorial nature of the opponent's action space.
Instead, we propose to solve a relaxation of this problem, by considering snapshots only constrained by MDP$_{t_0}$.
Making this approximation leaves a possibility to violate property \ref{prop:set-admissible-future-snapshot} but allows for an efficient search within the developed tree and (as will be shown experimentally) leads to robust policies.
For that purpose, we define the set of admissible snapshot models \wrt{} MDP$_{t_0}$ by $\Delta_{t_0}^t \eqdef \B_{W_1} \left( p_{t_0}(\cdot \mid s, a), L_p |t - t_0| \right) \times \B_{| \cdot |} \left( R_{t_0}(s, a), L_R |t - t_0| \right)$.
The relaxed analogues of Equations~\ref{eq:worst-case-v} and~\ref{eq:worst-case-q} for $s, t, a \in \S \times \T \times \A$ are defined as follows:
\begin{gather*}
	\hat{V}^{\pi}_{t_0,t}(s) \eqdef
	\min_{\left( p_i, R_i \right) \in \Delta_{t_0}^i, \forall i \in \T}
	\E \left[ \sum_{i = t}^{\infty} \gamma^{i-t} R_{i}(s_i, a_i)
	\Big|
	\begin{aligned}
		& s_t = s\\
		& a_i \sim \pi_i(\cdot \mid s_i), s_{i+1} \sim p_{i}(\cdot \mid s_i, a_i)
	\end{aligned} \right],\\
	\hat{Q}^\pi_{t_0,t}(s,a) \eqdef 
	\min_{\left({p},R\right) \in \Delta_{t_0}^t} \E_{s'\sim p}\left[R(s,a) + \gamma \hat{V}^{\pi}_{t_0, t + 1}(s') \right].
\end{gather*}
Their optimal counterparts, while seeking to find the optimal policy, verify the following equations:
\begin{align}
	\hat{V}^*_{t_0,t}(s) & = \max \limits_{a\in \A} \hat{Q}^*_{t_0,t}(s,a),
	\label{eq:relaxed-optimal-worst-case-v}\\
	\hat{Q}^*_{t_0,t}(s,a) & = \min \limits_{\substack{(p, R) \in \Delta_{t_0}^t}} \E\limits_{\substack{s' \sim p}} \left[ R(s, a) + \gamma \hat{V}^*_{t_0,t+1}(s') \right].
	\label{eq:relaxed-optimal-worst-case-q}
\end{align}
We now provide a method to calculate those quantities within the nodes of the tree search algorithm.\\
\textbf{Max nodes.}
A decision node $\nu^{s, t}$ corresponds to a max node due to the greediness of the agent \wrt{} the subsequent values of the children.
We aim at maximizing the return while retaining a risk-averse behavior.
As a result, the value of $\nu^{s, t}$ follows Equation~\ref{eq:relaxed-optimal-worst-case-v} and is defined as:
\begin{equation}
	\label{eq:decision-node-value}
	V(\nu^{s, t}) = \max_{a \in \A} V(\nu^{s, t, a}).
\end{equation}
\textbf{Min nodes.}
A chance node $\nu^{s, t, a}$ corresponds to a min node due to the use of a worst-case \NSMDP{} as a model which minimizes the value of $\nu^{s, t, a}$ \wrt{} the reward and the subsequent values of its children.
Writing the value of $\nu^{s, t, a}$ as the value of $s, t, a$, within the worst-case snapshot minimizing Equation~\ref{eq:relaxed-optimal-worst-case-q},  and using the children's values as values for the next reachable states, leads to Equation~\ref{eq:chance-node-value}.
\begin{equation}
	\label{eq:chance-node-value}
	V (\nu^{s, t, a}) =
	\min_{(p, R) \in \Delta_{t_0}^t}
	R(s, a) + \gamma
	\E_{s' \sim p}
	V (\nu^{s', t + 1})
\end{equation}
Our approach considers the environment as an adversarial agent, as in an \emph{asymmetric} two-player game, in order to search for a robust plan.
The resulting algorithm, \RATS{} for \RATSmeaning{}, is described in Algorithm~\ref{alg:rats}.
Given an initial state / decision epoch pair, a minimax tree is built using the snapshot MDP$_{t_0}$ and the operators corresponding to Equations~\ref{eq:decision-node-value} and~\ref{eq:chance-node-value} in order to estimate the worst-case snapshots at each depth.
The tree is built, the action leading to the best possible value from the root node is selected and a real transition is performed.
The next state is then reached, the new snapshot model MDP$_{t_0 + 1}$ is acquired and the process re-starts. 
Notice the use of $R(\nu)$ and $p(\nu' \mid \nu)$ in the pseudo-code: they are light notations respectively standing for $R_t(s, a)$ corresponding to a chance node $\nu \equiv \nu^{s, t, a}$ and the probability $p_t(s' | s, a)$ to jump to a decision node $\nu'\equiv \nu^{s', t + 1}$ given a chance node $\nu \equiv \nu^{s,t,a}$.
\begin{algorithm2e}[t]
	\DontPrintSemicolon
	\textbf{RATS} ($s_0$, $t_0$, maxDepth)\;
	$\nu_0$ = rootNode($s_0, t_0$)\;
	Minimax($\nu_0$)\;
	$\nu^* = \argmax_{\nu' \textbf{ in } \nu_0 \text{.children}} \nu'\text{.value}$\;
	\textbf{return }$\nu^*.\text{action}$\;
	\;
	\textbf{Minimax} ($\nu$, maxDepth)\;
	\If{$\nu$ is DecisionNode}{
		\If{$\nu$.{\normalfont state} is terminal {\normalfont \textbf{or}} $\nu$.{\normalfont depth} = {\normalfont maxDepth}}{
			\textbf{return }$\nu$.value = heuristicValue($\nu$.state)\;
		}
		\Else{
			\textbf{return }$\nu$.value = $\max_{\nu' \in \nu \text{.children}}${Minimax}($\nu'$, maxDepth)\;
		}
	}
	\Else{
		\textbf{return }$\nu\text{.value} = \min_{(p, R) \in \Delta_{t_0}^t} R(\nu) + \gamma \sum_{\nu' \in \nu\text{.children}} p(\nu' \mid \nu) \text{Minimax}(\nu', \text{maxDepth})$\;
	}
	\caption{RATS algorithm}
	\label{alg:rats}
\end{algorithm2e}
The tree built by \RATS{} is entirely developed until the maximum depth  $d_{\max}$.
A heuristic function is used to evaluate the leaf nodes of the tree.

\textbf{Analysis of \RATS{}.}
We are interested in characterizing Algorithm~\ref{alg:rats} without function approximation and therefore will consider finite, countable, $\S \times \A$ sets.
We now detail the computation of the min operator (Property~\ref{prop:close-form-worst-case-snapshot}), the computational complexity of \RATS{} (Property~\ref{prop:computational-complexity}) and the heuristic function.
\begin{property}
	\label{prop:close-form-worst-case-snapshot}
	\textbf{Closed-form expression of the worst case snapshot of a chance node.}
	Following Algorithm~\ref{alg:rats}, a solution to Equation~\ref{eq:chance-node-value} is given by:
	\begin{equation*}
		\hat{R}(s, a) = R_{t_0}(s, a) - L_R |t - t_0| \quad \textrm{and} \quad 
		\hat{p}(\cdot \mid s, a) = (1 - \lambda) p_{t_0}(\cdot \mid s, a) + \lambda p_{\text{sat}}(\cdot \mid s, a)
	\end{equation*}
	with $p_{\text{sat}}( \cdot \mid s, a) = (0, \cdots, 0, 1, 0, \cdots, 0)$ with $1$ at position $\argmin_{s'} V(\nu^{s', t + 1})$, $\lambda = 1 \text{ if } W_1(p_{\text{sat}}, p_0) \leq L_p |t - t_0|$ and $\lambda = L_p |t - t_0| / W_1 (p_{\text{sat}}, p_0)$ otherwise.
\end{property}
\begin{property}
	\textbf{Computational complexity.}
	The total computation complexity of Algorithm~\ref{alg:rats} is
	$\mathcal{O} (B |\S|^{1.5} |\A| \left( |\S| |\A| \right)^{d_{\max}})$
	with $B$ the number of time steps and $d_{\max}$ the maximum depth.
	\label{prop:computational-complexity}
\end{property}
\textbf{Heuristic function.}
As in vanilla minimax algorithms, Algorithm~\ref{alg:rats} bootstraps the values of the leaf nodes with a heuristic function if these leaves do not correspond to terminal states.
Given such a leaf node $\nu^{s, t}$, a heuristic aims at estimating 
the value of the optimal policy at $(s,t)$ within the worst-case \NSMDP{}, \ie{} $\hat{V}^*_{t_0,t}(s)$. 
Let $H(s,t)$ be such a heuristic function,
we call \emph{heuristic error} in $(s,t)$ the difference between $H(s,t)$ and $\hat{V}^*_{t_0,t}(s)$.
Assuming that the heuristic error is uniformly bounded, the following property provides an upper bound on the propagated error due to the choice of $H$.
\begin{property}
	\label{prop:heuristic-propagated-error}
	\textbf{Upper bound on the propagated heuristic error within \RATS{}.}
	Consider an agent executing Algorithm~\ref{alg:rats} at $s_0, t_0$ with a heuristic function $H$.
	We note $\mathcal{L}$ the set of all leaf nodes.
	Suppose that the heuristic error is uniformly bounded, \ie{}
	$
	\exists \delta > 0, \,
	\forall \nu^{s, t} \in \mathcal{L}, \,
	|H(s) - \hat{V}^*_{t_0,t}(s)| \leq \delta
	$.
	Then we have for every decision and chance nodes $\nu^{s, t}$ and $\nu^{s, t, a}$, at any depth $d \in [0, d_{\max}]$:
	\begin{align*}
		|V(\nu^{s, t}) - \hat{V}^*_{t_0,t}(s)|
		\leq \gamma^{(d_{\max} - d)} \delta
		\quad \text{ and } \quad
		|V(\nu^{s, t, a}) - \hat{Q}^*_{t_0,t}(s,a)|
		\leq \gamma^{(d_{\max} - d)} \delta.
	\end{align*}
\end{property}
This last result implies that with \emph{any} heuristic function $H$ inducing a uniform heuristic error, the propagated error at the root of the tree is guaranteed to be upper bounded by $\gamma^{d_{\max}} \delta$.
In particular, since the reward function is bounded by hypothesis, we have $\hat{V}^*_{t_0,t}(s) \leq 1 / (1 - \gamma)$.
Thus, selecting for instance the zero function ensures a root node heuristic error of at most ${\gamma^{d_{\max}}}/{(1 - \gamma)}$.
In order to improve the precision of the algorithm, we propose to guide the heuristic by using a function reflecting better the value of state $s$ at leaf node $\nu^{s, t}$.
The ideal function would of course be $H(s) = \hat{V}^*_{t_0,t}(s)$, reducing the heuristic error to zero, but this is intractable.
Instead, we suggest to  use the value of $s$ within the snapshot MDP$_t$ using an evaluation policy $\pi$, \ie{} $H(s) = V^\pi_{\text{MDP}_t}(s)$.
This snapshot is also not available, but Property~\ref{prop:bound-leaf-value} provides a range wherein this value lies. 
\begin{property}
	\label{prop:bound-leaf-value}
	\textbf{Bounds on the snapshots values.}
	Let $s \in \S$, $\pi$ a stationary  policy, MDP$_{t_0}$ and MDP$_t$ two snapshot MDPs, $t, t_0 \in \T^2$ be.
	We note $V^\pi_{\text{MDP}_{i}}(s)$ the value of $s$ within MDP$_{i}$ following $\pi$.
	Then,
	\begin{equation*}
	| V^\pi_{\text{MDP}_{t_0}}(s) - V^\pi_{\text{MDP}_t}(s) |
	\leq
	|t - t_0| L_R / (1-\gamma).
	\end{equation*}
\end{property}
Since $\text{MDP}_{t_0}$ is available, $V^\pi_{\text{MDP}_{t_0}}(s)$ can be estimated, \eg{} via Monte-Carlo roll-outs. Let $\widehat{V}^\pi_{\text{MDP}_{t_0}}(s)$ denote such an estimate.
Following Property~\ref{prop:bound-leaf-value}, $V^\pi_{\text{MDP}_{t_0}}(s) - |t - t_0| L_R / (1-\gamma) \leq V^\pi_{\text{MDP}_t}(s)$.
Hence, a worst-case heuristic on $V^\pi_{\text{MDP}_t}(s)$ is $H(s) = \widehat{V}^\pi_{\text{MDP}_{t_0}}(s) - |t - t_0| L_R / (1-\gamma)$.
The bounds provided by Property~\ref{prop:heuristic-propagated-error} decrease quickly with $d_{\max}$, and given that $d_{\max}$ is \emph{large enough}, 
\RATS{} provides the optimal risk-averse maximizing the worst-case value for any evolution of the \NSMDP{}.

\section{Experiments}
\label{sec:expe}

We compare the \RATS{} algorithm with two policies~\footnote{
	\if\statecodewillbeprovided1
	Code: https://github.com/SuReLI/rats-experiments
	\fi
	--
	ML reproducibility checklist: Appendix Section~\ref{sec:reproducibility-checklist}.
}.
The first one, named DP-snapshot, uses Dynamic Programming to compute the optimal actions \wrt{} the snapshot models at each decision epoch.
The second one, named DP-NSMDP, uses the real NSMDP as a model to provide its optimal action.
The latter behaves as an omniscient agent and should be seen as an upper bound on the performance.
We choose a particular grid-world domain coined ``Non-Stationary bridge'' illustrated in Appendix, Section~\ref{sec:nsbridge-figure}.
An agent starts at the state labeled S in the center and the goal is to reach one of the two terminal states labeled G where a reward of +1 is received.
The gray cells represent holes that are terminal states where a reward of -1 is received.
Reaching the goal on the right leads to the highest payoff since it is closest to the initial state and a discount factor $\gamma = 0.9$ is applied.
The actions are $\A = \{ \text{Up, Right, Down, Left} \}$.
The transition function is stochastic and non-stationary.
At decision epoch $t = 0$, any action deterministically yields the intuitive outcome.
With time, when applying Left or Right, the probability to reach the positions usually stemming from Up and Down increases symmetrically until reaching 0.45.
We set the Lipschitz constant $L_p = 1$.
Aside, we introduce a parameter $\epsilon \in [0, 1]$ controlling the behavior of the environment.
If $\epsilon = 0$, only the left-hand side bridge becomes slippery with time. It reflects a close to worst-case evolution for a policy aiming to the left-hand side goal.
If $\epsilon = 1$, only the right-hand side bridge becomes slippery with time. It reflects a close to worst-case evolution for a policy aiming to the right-hand side goal.
In between, the misstep probability is proportionally balanced between left and right.
One should note that changing $\epsilon$ from 0 to 1 does not cover all the possible evolutions from MDP$_{t_0}$ but provides a concrete, graphical illustration of \RATS's behavior for various possible evolutions of the \NSMDP{}.

We tested \RATS{} with $d_{\max} = 6$ so that leaf nodes in the search tree are terminal states. Hence, the optimal risk-averse policy is applied and no heuristic approximation is made.
Our goal is to demonstrate that planning in this worst-case \NSMDP{} allows to minimize the loss given any possible evolution of the environment.
To illustrate this, we report results reflecting different evolutions of the same \NSMDP{} using the $\epsilon$ factor.
It should be noted that, at $t = 0$, \RATS{} always moves to the left, even if the goal is further, since going to the right may be risky if the probabilities to go Up and Down increase.
This corresponds to the careful, risk-averse, behavior.
Conversely, DP-snapshot always moves to the right since $\text{MDP}_{0}$ does not capture this risk.
As a result, the $\epsilon = 0$ case reflects a favorable evolution for DP-snapshot and a bad one for \RATS{}.
The opposite occurs with $\epsilon = 1$ where the cautious behavior dominates over the risky one, and the in-between cases mitigate this effect.

\begin{figure}[]
	\begin{subfigure}{.5\textwidth}
		\centerline{
		\includegraphics[trim={0cm, 0cm, 0cm, 0mm}, clip, width=\textwidth]{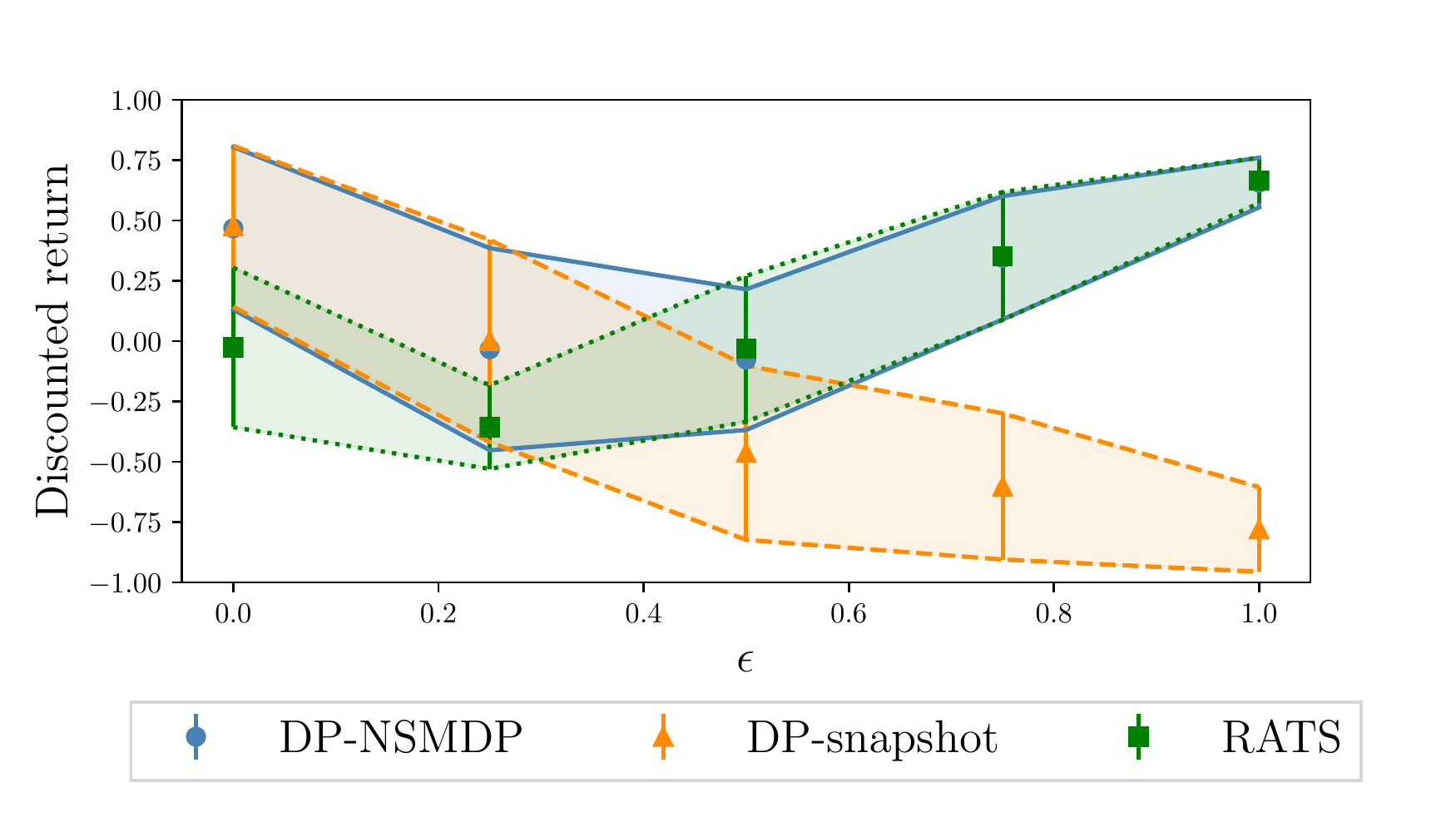}}
		\captionsetup{width=\textwidth}
		\caption{Discounted return vs $\epsilon$, 50\% of standard deviation.}
		\label{fig:return-vs-epsilon}
	\end{subfigure}%
	\begin{subfigure}{.5\textwidth}
		\centerline{
		\includegraphics[trim={2cm, 3mm, 2cm, 5mm}, clip, width=\textwidth,height=10.5em]{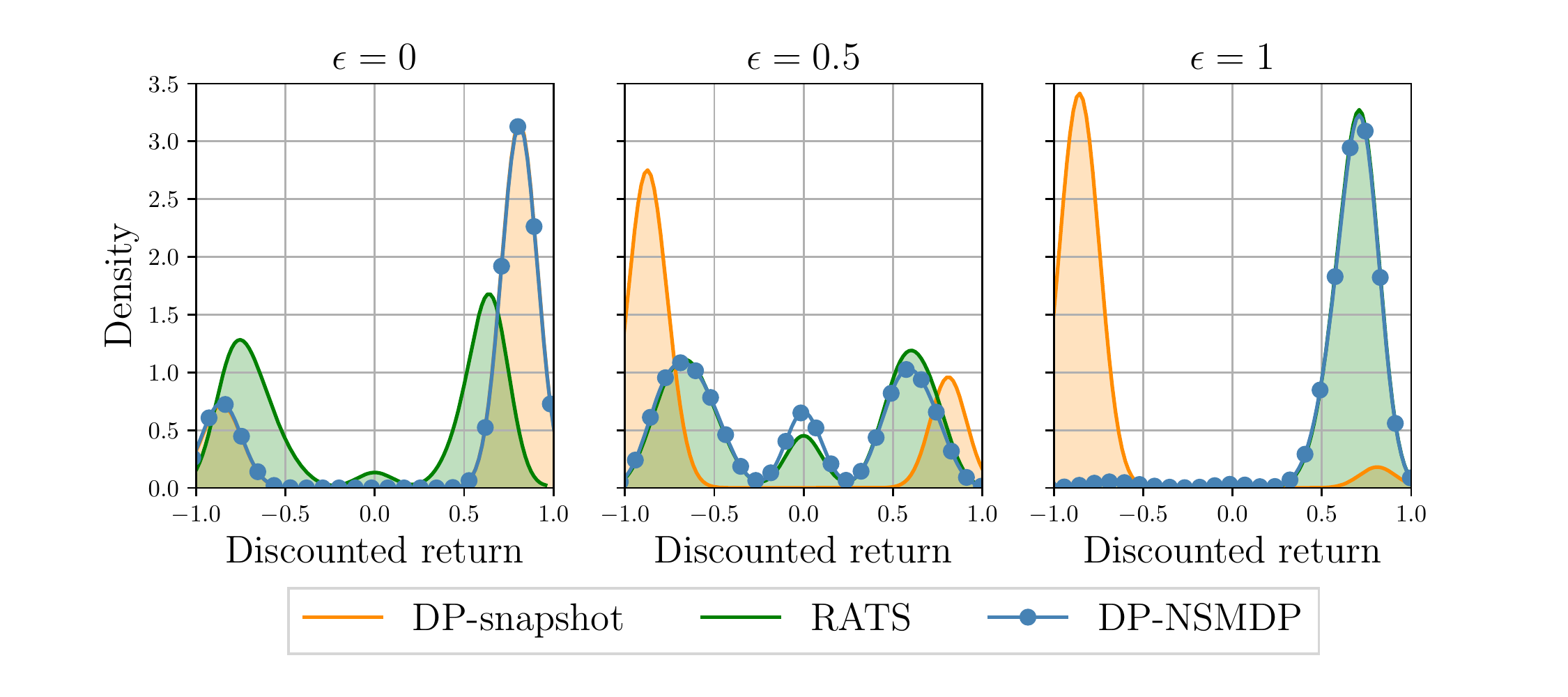}}
		\captionsetup{width=0.9\textwidth}
		\caption{Discounted return distributions $\epsilon \in \{ 0, 0.5, 1 \}$.}
		\label{fig:return-distributions}
	\end{subfigure}
	\caption{Discounted return of the three algorithms for various values of $\epsilon$.}
\end{figure}

In Figure~\ref{fig:return-vs-epsilon}, we display the achieved expected return for each algorithm as a function of $\epsilon$, \ie{} as a function of the possible evolutions of the \NSMDP{}.
As expected, the performance of DP-snapshot strongly depends on this evolution. It achieves high return for $\epsilon = 0$ and low return for $\epsilon = 1$.
Conversely, the performance of \RATS{} varies less across the different values of $\epsilon$.
The effect illustrated here is that \RATS{} maximizes the minimal possible return given \emph{any} evolution of the \NSMDP{}.
It provides the guarantee to achieve the best return in the worst-case. This behavior is highly desirable when one requires robust performance guarantees as, for instance, in critical certification processes.
%
%
%
Figure~\ref{fig:return-distributions} displays the return distributions of the three algorithms for $\epsilon \in \{ 0, 0.5, 1 \}$.
The effect seen here is the tendency for \RATS{} to diminish the left tail of the distribution corresponding to low returns for each evolution.
It corresponds to the optimized criteria, \ie{} robustly maximizing the worst-case value.
A common risk measure is the Conditional Value at Risk (CVaR) defined as the expected return in the worst $q$\% cases.
We illustrate the CVaR at 5\% achieved by each algorithm in Table~\ref{table:result}.
Notice that \RATS{} always maximizes the CVaR compared to both DP-snapshot and DP-\NSMDP{}.
Indeed, even if the latter uses the true model, the optimized criteria in DP is the expected return.

\section{Conclusion}
\label{sec:conclu}

We proposed an approach for robust planning in non-stationary stochastic environments.
We introduced the framework of Lipchitz Continuous Non-Stationary MDPs (\NSMDP{}s) and derived the Risk-Averse Tree-Search (\RATS{}) algorithm, to predict the worst-case evolution and to plan optimally \wrt{} this worst-case \NSMDP{}.
We analyzed \RATS{} theoretically and showed that it approximates a worst-case \NSMDP{} with a control parameter that is the depth of the search tree.
We showed empirically the benefit of the approach that searches for the highest lower bound on the worst achievable score.
\RATS{} is robust to every possible evolution of the environment, \ie{} maximizing the expected worst-case outcome on the whole set of possible \NSMDP{}s.
Our method was applied to the uncertainty on the evolution of a model.
Generally, it could be extended to any uncertainty on the model used for planning, given bounds on the set of the feasible models.
The purpose of this contribution is to lay a basis of worst-case analysis for robust solutions to \NSMDP{}s.
As is, \RATS{} is computationally intensive and scaling the algorithm to larger problems is an exciting future challenge.

\section*{Acknowledgments}

This research was supported by the Occitanie region, France.

\bibliographystyle{abbrvnat}
\bibliography{ratsbib}

\if\includeappendixatbottom1


\section*{Supplementary material}

# Non-Stationary Markov Decision Processes a Worst-Case Approach using Model-Based Reinforcement Learning

# Appendix

**Erwan Lecarpentier**
Université de Toulouse
ONERA - The French Aerospace Lab
erwan.lecarpentier@isae-supaero.fr

**Emmanuel Rachelson**
Université de Toulouse
ISAE-SUPAERO
emmanuel.rachelson@isae-supaero.fr

In the following proofs, the dual formulation of the 1-Wasserstein distance is used several times. We include the definition here for reference purpose.

**Definition 1.** *Dual formulation of the 1-Wasserstein distance.* Let $(X, d_X)$ be a Polish metric space and $\mu, \nu$ any two probability measures on $X$. The dual formulation of the 1-Wasserstein distance between $\mu$ and $\nu$ is defined by

$$W_1(\mu, \nu) = \sup_{f \in Lip_1} \int_X f(x) d(\mu - \nu)(x) \qquad (1)$$

where $Lip_1$ denotes the set of the continuous mappings $X \to \mathbb{R}$ with a minimal Lipschitz constant bounded by 1.

## 1  Proof of Property 1

Consider an $(L_p, L_r)$-LC-NSMDP. Let $s, t, a, \hat{t} \in \mathcal{S} \times \mathcal{T} \times \mathcal{A} \times \mathcal{T}$ be. By definition of the expected reward function, the following holds:

$$\begin{aligned}
R_t(s,a) - R_{\hat{t}}(s,a) &= \int_\mathcal{S} \Big( p_t(s' \mid s,a) r_t(s,a,s') - p_{\hat{t}}(s' \mid s,a) r_{\hat{t}}(s,a,s') \Big) ds' \\
&= \int_\mathcal{S} \Big( r_t(s,a,s') \big[ p_t(s' \mid s,a) - p_{\hat{t}}(s' \mid s,a) \big] \\
&\quad + p_{\hat{t}}(s' \mid s,a) \big[ r_t(s,a,s') - r_{\hat{t}}(s,a,s') \big] \Big) ds' \\
&= \int_\mathcal{S} r_t(s,a,s') \big[ p_t(s' \mid s,a) - p_{\hat{t}}(s' \mid s,a) \big] ds' \\
&\quad + \int_\mathcal{S} p_{\hat{t}}(s' \mid s,a) \big[ r_t(s,a,s') - r_{\hat{t}}(s,a,s') \big] ds' \\
&\leq \sup_{\|f\|_L \leq 1} \int_\mathcal{S} f(s', t') \big[ p_t(s' \mid s,a) - p_{\hat{t}}(s' \mid s,a) \big] ds' \\
&\quad + \int_\mathcal{S} p_{\hat{t}}(s' \mid s,a) L_r |t - \hat{t}| ds' \\
&\leq W_1(p(\cdot \mid s,t,a), p(\cdot \mid s,\hat{t},a)) + L_r |t - \hat{t}| \\
&\leq (L_p + L_r)|t - \hat{t}|
\end{aligned}$$

Where we used the triangle inequality, the fact that $r$ is a bounded function and the dual formulation of the 1-Wasserstein distance (see Definition 1). The same inequality can be derived with the opposite terms which concludes the proof by taking the absolute value.

## 2 Proof of Property 2

*Proof.* The proof is straightforward using the Lipschitz property of Definition 4 and Property 1. □

## 3 Proof of Property 4

Let us first calculate the cost of constructing a tree with the minimax procedure. Following Algorithm 1, a tree is composed of at most $n_l$ leaf nodes, $n_d$ non-leaf decision nodes and $n_c$ chance nodes, with the following values for the integers $n_l$, $n_d$ and $n_c$:

$$n_l = (|\mathcal{S}||\mathcal{A}|)^{d_{\max}}, n_d = \sum_{i=0}^{d_{\max}-1} (|\mathcal{S}||\mathcal{A}|)^i \text{, and } n_c = |\mathcal{A}|B.$$

As a result, we have that $n_l$ is $\mathcal{O}((|\mathcal{S}||\mathcal{A}|)^{d_{\max}})$, $n_d$ is $\mathcal{O}((|\mathcal{S}||\mathcal{A}|)^{d_{\max}-1})$ and $n_c$ is $\mathcal{O}(|\mathcal{A}|(|\mathcal{S}||\mathcal{A}|)^{d_{\max}-1})$. We note respectively $c_l$, $c_d$ and $c_c$ the number of operations required to compute the values of a leaf node, a non-leaf decision node and a chance node. To compute the whole tree we need to build *and* evaluate all the nodes, resulting in at most the following number of operations:

$$n_l c_l \times n_d c_d \times n_c c_c. \tag{2}$$

We will assume that $c_l$ is $\mathcal{O}(1)$ without further details on the nature of the heuristic function. As the value of a non-leaf decision node is computed by finding the maximum value among the $|\mathcal{A}|$ children, we have that $c_d$ is $\mathcal{O}(|\mathcal{A}|)$. From Theorem 3, the evaluation of a chance node is equivalent to computing a 1-Wasserstein distance, which is a linear program. Following Vaidya's algorithm [Vaidya, 1989], the cost in the worst-case is $\mathcal{O}(|\mathcal{S}|^{2.5})$ where $|\mathcal{S}|$ is the dimension of the problem in our case. As a result, $c_c$ is $\mathcal{O}(|\mathcal{S}|^{2.5})$. Replacing all the values in Equation 2, we deduce that the total number of operation of computing a tree is

$$\mathcal{O}\left(|\mathcal{S}|^{1.5}\left(|\mathcal{S}||\mathcal{A}|\right)^{d_{\max}}\right).$$

After computing a tree, the action maximizing the value should be selected which has complexity $\mathcal{O}(|\mathcal{A}|)$. The operation being repeated for every time steps, one should multiply everything by $B$, the total number of time steps for which the algorithm is run. As a result, the total computational complexity of RATS is

$$\mathcal{O}\left(B|\mathcal{S}|^{1.5}|\mathcal{A}|\left(|\mathcal{S}||\mathcal{A}|\right)^{d_{\max}}\right).$$

## 4 Proof of Property 3

We are looking for a closed-form expression of the value of a chance node $\nu^{s,t,a}$ as defined in Equation 6 recalled below.

$$(\bar{p}, \bar{R}) = \underset{(p,R) \in \Delta_{t_0,t}}{\arg\min} R(s,a) + \gamma \mathbb{E}_{s' \sim p(\cdot|s,a)} V(\nu^{s',t+1})$$

Obviously, we have that $\bar{R} = R_{t_0}(s,a) - L_R|t - t_0|$ and $\bar{p}$ is given by:

$$\bar{p} = \underset{p \in \mathcal{B}_{W_1}(p_{t_0}(\cdot|s,a), L_p|t-t_0|)}{\arg\min} \sum_{s'} p(s' \mid s,a) V(\nu^{s',t+1})$$

where $\mathcal{B}_d(c, r)$ denotes the ball of center $c$, defined with metric $d$ and radius $r$. Since we are in the discrete case, we enumerate through the elements of $\mathcal{S}$ and write the vectors $p \equiv (p(s' \mid s, a))_{s'}$,

2

$p_0 \equiv (p_{t_0}(s' \mid s, a))_{s'}$ and $v \equiv (V(\nu^{s',t+1}))_{s'}$. The problem can then be re-written as follows:

$$\bar{p} = \arg\min_p \quad p^\top v \tag{3}$$

$$\text{s.t. } p^\top \mathbf{1} = 1 \tag{4}$$

$$p \geq 0 \tag{5}$$

$$W_1(p, p_0) \leq C \tag{6}$$

Where we have $\mathbf{1} \in \mathbb{R}^{|\mathcal{S}|}$ a vector of ones, $C = L_p |t - t_0|$ and the 1-Wasserstein metric between two discrete distributions written in dual form following Lemma 1 as:

$$W_1(u, v) = \max_f f^\top (u - v) \tag{7}$$

$$\text{s.t. } Af \leq b$$

Where the matrix $A$ and vector $b$ are defined such that for any indexes $i, j$ we have $|f_i - f_j| \leq d_{i,j}$ with $d_{i,j}$ the metric defined over the measured space, in our case the state space $\mathcal{S}$. Hence we propose to solve the program 3 under constraints 4 to 6. Let us first show that this problem is convex. Clearly, the objective function in Equation 3 is linear, hence convex, and the constraints 4 and 5 define a convex set. We prove that the 1-Wasserstein distance is convex in Lemma 1.

**Lemma 1.** *Convexity of the 1-Wasserstein distance. The 1-Wasserstein distance is convex i.e. for $\lambda \in [0, 1]$, $(X, d_X)$ a Polish space and any three probability measures $w_0, w_1, w_2$ on $X$, the following holds:*

$$W_1(w_0, \lambda w_1 + (1 - \lambda)w_2) \leq \lambda W_1(w_0, w_1) + (1 - \lambda)W_1(w_0, w_2)$$

*Proof.* We use the dual representation of the 1-Wasserstein distance of Definition 1.

$$W_1(w_0, \lambda w_1 + (1 - \lambda)w_2)$$

$$= \sup_{f \in \text{Lip}_1} \int_X f(x)(w_0(x) - \lambda w_1(x) - (1 - \lambda)w_2(x))dx$$

$$= \sup_{f \in \text{Lip}_1} \int_X \left(\lambda f(x)(w_0(x) - w_1(x)) + (1 - \lambda)f(x)(w_0(x) - w_2(x))\right) dx$$

$$\leq \lambda \sup_{f \in \text{Lip}_1} \int_X f(x)(w_0(x) - w_1(x))dx + (1 - \lambda) \sup_{f \in \text{Lip}_1} \int_X f(x)(w_0(x) - w_2(x))dx$$

$$\leq \lambda W_1(w_0, w_1) + (1 - \lambda)W_1(w_0, w_2)$$

Where we used the linearity of the integral and the triangle inequality on the sup operator. □

The program 3 is thus convex. One can also observe that the gradient of the objective function is constant, equal to $+v$. Furthermore, $p_0$ is an admissible initial point that we could use for a gradient descent method. However, given $p_0$, following the descent direction $-v$ may break the constraints 4 and 5. One would have to project this gradient onto a certain, unknown, set of hyperplanes in order to apply the gradient method descent. Let us note $\text{proj}(v)$ the resulting projected gradient, that is unknown.

We remark that the vector $p_{\text{sat}} = (0, \cdots, 0, 1, 0, \cdots, 0)$ with 1 at the index $\arg\min_i v_i$ where $v_i$ denotes the $i$th coefficient of $v$, is the optimal solution of the program 3 when we remove the Wasserstein constraint 6. One can observe that the optimal solution with the constraint 6 would as well be $p_{\text{sat}}$ if the constant $C$ is *big enough*. As a result, the descent direction $\nabla = p_{\text{sat}} - p_0$ is the one to be followed in this setting when applying the gradient descent method to this case. Furthermore, following $\nabla$ from $p_0$ until $p_{\text{sat}}$ never breaks the constraints 4 and 5. Since the gradient of the objective function is constant, there can exist only one $\text{proj}(v)$. $\nabla$ fulfils the requirements, hence we have $\text{proj}(v) = \nabla$.

We can now apply the gradient method descent with the following 1-shot rule since the gradient is constant:

$$\bar{p} := p_0 + \lambda \nabla \text{ with, } \begin{cases} \lambda = 1 \text{ if } W_1(p_{\text{sat}}, p_0) \leq C \\ \lambda = C/W_1(p_{\text{sat}}, p_0) \end{cases}$$

Indeed, in the first case, we can follow $\nabla$ until the extreme distribution $p_{\text{sat}}$ without breaking the constraint 6. Going further is trivially infeasible.

In the second case, we have to stop *in between* so that the constraint 6 is saturated. In such a case, we cannot go further without breaking this constraint and we recall that no projected gradient could be found by uniqueness of this gradient in our setting. Hence we have the following equality:

$$W_1(p_0 + \lambda\nabla, p_0) = C$$
$$\max_{Af \leq b} f^\top (p_0 + \lambda\nabla - p_0) = C$$
$$\lambda \max_{Af \leq b} f^\top \nabla = C$$
$$\lambda = C/W_1(p_{\text{sat}}, p_0)$$

Where we used the fact that $\nabla = p_{\text{sat}} - p_0$. The latter result concludes the proof.

## 5 Proof of Property 5

Let us consider a tree developed with Algorithm 1 with a heuristic function $H : s \mapsto H(s)$ used to estimate the value of a leaf node. The set of the leaves nodes is denoted by $\mathcal{L}$ and we have the following uniform upper bound $\delta > 0$ on the heuristic error:

$$\forall \nu^{s,t} \in \mathcal{L}, \ |H(s) - \overline{V}^*_{t_0,t}(s)| < \delta \tag{8}$$

We want to prove the following result for a decision and chance nodes $\nu^{s,t}$ and $\nu^{s,t,a}$ at any depth $d \in [0, d_{\max}]$:

$$|V(\nu^{s,t}) - \overline{V}^*_{t_0,t}(s)| \leq \gamma^{(d_{\max}-d)} \delta \tag{9}$$
$$|V(\nu^{s,t,a}) - \overline{Q}^*_{t_0,t}(s,a)| \leq \gamma^{(d_{\max}-d)} \delta \tag{10}$$

The proof is made by induction, starting at depth $d_{\max}$ and reversely ending at depth 0. At $d_{\max}$, the nodes are leaf nodes, their values is estimated with the heuristic function i.e. $V(\nu^{s,t}) = H(s)$. Hence the result is directly proven by hypothesis in Equation 8. We will now start by proving the result for the chance nodes which come as the first parents of the decision node for which we initialized the induction proof. Then we extend it to the parents decision nodes which completes the proof.

**Chance nodes case.** Consider any chance node $\nu^{s,t,a}$ at depth $d \in [0, d_{\max}]$. We suppose that the property is true for depth $d+1$, thus we have for any decision node at $d+1$ denoted by $\nu^{s',t'}$:

$$|V(\nu^{s',t'}) - \overline{V}^*_{t_0,t'}(s')| \leq \gamma^{(d_{\max}-(d+1))} \delta$$

Following Equation 6 of the paper, we have by construction:

$$V(\nu^{s,t,a}) = \overline{R}_t(s,a) + \gamma \sum_{s'} \overline{p}_t(s' \mid s,a) V(\nu^{s',t'})$$

By definition, the true $Q$-value function defined by the Bellman Equation 2 gives the true target value:

$$\overline{Q}^*_{t_0,t}(s,a) = \overline{R}_t(s,a) + \gamma \sum_{s'} \overline{p}_t(s' \mid s,a) \overline{V}^*_{t_0,t'}(s')$$

Hence, using the induction hypothesis, we have the following inequalities proving the result of Equation 10:

$$|V(\nu^{s,t,a}) - \overline{Q}^*_{t_0,t}(s,a)| = \gamma \left| \sum_{s'} \overline{p}_t(s' \mid s,a) V(\nu^{s',t'}) - \sum_{s'} \overline{p}_t(s' \mid s,a) \overline{V}^*_{t_0,t'}(s') \right|$$
$$\leq \gamma \sum_{s'} \overline{p}_t(s' \mid s,a) \left| V(\nu^{s'}) - \overline{V}^*_{t_0,t'}(s') \right|$$
$$\leq \gamma \sum_{s'} \overline{p}_t(s' \mid s,a) \gamma^{(d_{\max}-(d+1))} \delta$$
$$\leq \gamma^{(d_{\max}-d)} \delta$$

**Decision nodes case.** Consider now any decision node $\nu^{s,t}$ at the same depth $d \in [0, d_{\max})$. The value of such a node is given by Equation 5 of the paper and the following holds.
$$V(\nu^{s,t}) = V(\nu^{s,t,\bar{a}}), \text{ with, } \bar{a} = \arg\max_{a \in \mathcal{A}} V(\nu^{s,t,a})$$

Similarly, we define $a^* \in \mathcal{A}$ as follows:
$$\overline{V}^*_{t_0,t}(s) = \overline{Q}^*_{t_0,t}(s, a^*), \text{ with, } a^* = \arg\max_{a \in \mathcal{A}} \overline{Q}^*_{t_0,t}(s, a)$$

We distinguish two cases: 1) if $\bar{a} = a^*$ and 2) if $\bar{a} \neq a^*$. In case 1), the result is trivial by writing the value of the decision node as the value of the chance node with the action $a^*$ and using the – already proven for depth $d$ – result of Equation 10.
$$|V(\nu^{s,t}) - \overline{V}^*_{t_0,t}(s)| = |V(\nu^{s,t,a^*}) - \overline{Q}^*_{t_0,t}(s, a^*)|$$
$$\leq \gamma^{(d_{\max}-d)}\delta$$

In case 2), the maximizing actions are different. Still following Equation 10, we have that $V(\nu^{s,t,a^*}) \geq \overline{Q}^*_{t_0,t}(s, a^*) - \gamma^{(d_{\max}-d)}\delta$. Yet, since $\bar{a}$ is the maximizing action in the tree, we have that $V(\nu^{s,t,\bar{a}}) \geq V(\nu^{s,t,a^*})$. By transitivity, we can thus write the following:
$$V(\nu^{s,t,\bar{a}}) \geq \overline{Q}^*_{t_0,t}(s, a^*) - \gamma^{(d_{\max}-d)}\delta$$
$$\Rightarrow \quad \overline{Q}^*_{t_0,t}(s, a^*) - V(\nu^{s,t,\bar{a}}) \leq \gamma^{(d_{\max}-d)}\delta \tag{11}$$

Furthermore, still following Equation 10, we have that $\overline{Q}^*_{t_0,t}(s, \bar{a}) \geq V(\nu^{s,t,\bar{a}}) - \gamma^{(d_{\max}-d)}\delta$. Yet, since $a^*$ is the maximizing action in $\widehat{\text{MDP}}$, we have that $\overline{Q}^*_{t_0,t}(s, a^*) \geq \overline{Q}^*_{t_0,t}(s, \bar{a})$. By transitivity, we can thus write the following:
$$\overline{Q}^*_{t_0,t}(s, a^*) \geq V(\nu^{s,t,\bar{a}}) - \gamma^{(d_{\max}-d)}\delta$$
$$\Rightarrow \quad V(\nu^{s,t,\bar{a}}) - \overline{Q}^*_{t_0,t}(s, a^*) \leq \gamma^{(d_{\max}-d)}\delta \tag{12}$$

By assembling equations 11 and 12, we prove equation 9 and the proof by induction is complete.

## 6 Proof of Property 6

Let $s, t_0, t \in \mathcal{S} \times \mathcal{T} \times \mathcal{T}$ be. We consider the two snapshots $\text{MDP}_{t_0}$ and $\text{MDP}_t$ and are interested in the values of $s$ within those two snapshots using the random policy $\pi$. We note $V^\pi_{\text{MDP}_{t_0}}(s)$ and $V^\pi_{\text{MDP}_t}(s)$ those values. Let $n \in \mathbb{N}$ be. We note $V^{\pi,n}_{\text{MDP}_{t_0}}(s)$ and $V^{\pi,n}_{\text{MDP}_t}(s)$ the finite horizon values defined as follows:

$$V^{\pi,n}_{\text{MDP}_{t_0}}(s) = \mathbb{E}\left\{\sum_{i=0}^n \gamma^i r_{t_0}(s_i, a_i, s_{i+i}) \middle| \begin{array}{l} s_0 = s, \\ s_{i+1} \sim p_{t_0}(\cdot \mid s_i, a_i), i \geq 0 \\ a_i \sim \pi(\cdot), i \geq 0 \end{array}\right\}$$

where we replace $t_0$ by $t$ for the definition of $V^{\pi,n}_{\text{MDP}_t}(s)$. We first prove a result on the finite horizon values in Lemma 2.

**Lemma 2.** *We consider an $(L_p, L_R)$-LC-NSMDP. For $s, t, t_0 \in \mathcal{S} \times \mathcal{T} \times \mathcal{T}$ and $n \in \mathbb{N}$, the finite horizon of the values of $s$ within the snapshots $\text{MDP}_t$ and $\text{MDP}_{t_0}$ verify:*
$$|V^{\pi,n}_{MDP_{t_0}}(s) - V^{\pi,n}_{MDP_t}(s)| \leq L_{V_n}|t - t_0|$$
$$\text{with, } L_{V_n} = \sum_{i=0}^n \gamma^i L_R$$

*Proof.* The proof is made by induction. Let us start with $n = 0$. By definition, we have:
$$\left|V^{\pi,0}_{\text{MDP}_{t_0}}(s) - V^{\pi,0}_{\text{MDP}_t}(s)\right| = \left|\int_{\mathcal{A}} \pi(a \mid s)\left(R_{t_0}(s, a) - R_t(s, a)\right) da\right|$$
$$\leq \int_{\mathcal{A}} \pi(a \mid s) L_R |t_0 - t| da$$
$$\leq L_R |t_0 - t|$$

5

Which verifies the property for $n = 0$ with $L_{V_0} = L_R$. Let us now consider $n \in \mathbb{N}$ and suppose the property true for rank $n - 1$. By writing the Bellman equation for the two value functions, we obtain the following calculation:

$$V_{\text{MDP}_{t_0}}^{\pi,n}(s) - V_{\text{MDP}_t}^{\pi,n}(s) = \int_{\mathcal{S} \times \mathcal{A}} \pi(a|s) \Big[ p_{t_0}(s' \mid s, a)(r_{t_0}(s, a, s') + \gamma V_{\text{MDP}_{t_0}}^{\pi,n-1}(s')) -$$

$$p_t(s' \mid s, a)(r_t(s, a, s') + \gamma V_{\text{MDP}_t}^{\pi,n-1}(s')) \Big] ds' da$$

i.e. $V_{\text{MDP}_{t_0}}^{\pi,n}(s) - V_{\text{MDP}_t}^{\pi,n}(s) = \int_{\mathcal{A}} \pi(a|s) \Big[ A(s,a) + B(s,a) \Big] da \qquad (13)$

With the following values for $A(s,a)$ and $B(s,a)$:

$$A(s,a) = \int_{\mathcal{S}} (r_{t_0}(s, a, s') + \gamma V_{\text{MDP}_{t_0}}^{\pi,n-1}(s')) \Big[ p_{t_0}(s' \mid s, a) - p_t(s' \mid s, a) \Big] ds'$$

$$B(s,a) = \int_{\mathcal{S}} p_t(s' \mid s, a) \Big[ r_{t_0}(s, a, s') - r_t(s, a, s') + \gamma (V_{\text{MDP}_{t_0}}^{\pi,n-1}(s') - V_{\text{MDP}_t}^{\pi,n-1}(s')) \Big] ds'$$

Let us first bound $A(s,a)$ by noticing that $s' \mapsto r_{t_0}(s, a, s') + \gamma V_{\text{MDP}_{t_0}}^{\pi,n-1}(s')$ is bounded by $\frac{1}{1-\gamma}$. Since the function $s' \mapsto \frac{1}{1-\gamma}$ belongs to $\text{Lip}_1$, we can write the following:

$$A(s,a) \leq \sup_{f \in \text{Lip}_1} \int_{\mathcal{S}} f(s') \Big[ p_{t_0}(s' \mid s, a) - p_t(s' \mid s, a) \Big] ds'$$

$$\leq W_1(p_{t_0}, p_t)$$

$$\leq L_p |t - t_0|$$

$B$ is straightforwardly bounded using the induction hypothesis:

$$B(s,a) \leq \int_{\mathcal{S}} p_t(s' \mid s, a) \Big[ L_r |t - t_0| + \gamma \sum_{i=0}^{n-1} \gamma^i L_R |t - t_0| \Big] ds'$$

$$\leq L_r |t - t_0| + \sum_{i=1}^{n} \gamma^i L_R |t - t_0|$$

We inject the result in Equation 13:

$$V_{\text{MDP}_{t_0}}^{\pi,n}(s) - V_{\text{MDP}_t}^{\pi,n}(s) \leq \int_{\mathcal{A}} \pi(a|s) \Big[ L_p |t - t_0| + L_r |t - t_0| + \sum_{i=1}^{n} \gamma^i L_R |t - t_0| \Big] da$$

$$\leq (L_p + L_r)|t - t_0| + \sum_{i=1}^{n} \gamma^i L_R |t - t_0|$$

$$\leq L_R |t - t_0| + \sum_{i=1}^{n} \gamma^i L_R |t - t_0|$$

$$\leq \sum_{i=0}^{n} \gamma^i L_R |t - t_0|$$

The same result can be derived with the opposite expression. Hence, taking the absolute value, we prove the property at rank $n$, i.e.

$$|V_{\text{MDP}_{t_0}}^{\pi,n}(s) - V_{\text{MDP}_t}^{\pi,n}(s)| \leq \sum_{i=0}^{n} \gamma^i L_R |t - t_0| \qquad (14)$$

which concludes the proof by induction. $\square$

The proof of Property 6 follows easily by remarking that the sequence $L_{V_n}$ of Lemma 2 is geometric and converges towards $\frac{L_R}{1-\gamma}$ when $n$ goes to infinity.

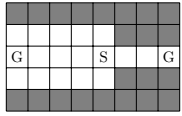

Figure 1: The Non-Stationary bridge environment

# 7 Non-Stationary bridge environment

# 8 Informations about the Machine Learning reproducibility checklist

For the experiments run in Section 6, the computing infrastructure used was a laptop using four 64-bit CPU (model: Intel(R) Core(TM) i7-4810MQ CPU @ 2.80GHz). The collected samples sizes and number of evaluation runs for each experiment are summarized in Table 1.

| Experiment | Number of experiment repetitions | Number of episodes | Maximum length of episodes | Upper bound on the number of computed transition samples $(s, a, r, s')$ |
|---|---|---|---|---|
| Non-Stationary Bridge Figure 1 | 3 (one per agent) | 96 | 10 | 89,579,520 |

Table 1: Summary of the number of experiment repetition, number of sampled tasks, number of episodes, maximum length of episodes and upper bounds on the number of collected samples.

The displayed confidence intervals in Figure 2a is 50% of the estimated confidence interval $\bar{\sigma}$ computed w.r.t. the following formula:

$$\bar{\sigma} = \sqrt{\frac{1}{1-N} \sum_{i=1}^{N} (x_i - \bar{x})^2} \quad \text{where,} \quad \bar{x} = \frac{1}{N} \sum_{i=1}^{N} x_i,$$

with $D = \{x_i\}_{i=1}^{N}$ the set of the collected data (discounted return in this case). No data were excluded neither pre-computed. Hyper-parameters were determined to our appreciation, they may be sub-optimal but we found the results convincing enough to display interesting behaviours.

# References

Pravin M. Vaidya. Speeding-up linear programming using fast matrix multiplication. In *30th Annual Symposium on Foundations of Computer Science*, pages 332–337. IEEE, 1989.


\fi

\end{document}